\pdfoutput=1

\documentclass[11pt]{article}

\usepackage[final]{acl}

\usepackage{times}
\usepackage{latexsym}

\usepackage[T1]{fontenc}

\usepackage[utf8]{inputenc}

\usepackage{microtype}

\usepackage{inconsolata}

\usepackage{graphicx}
\usepackage{url}
\usepackage{booktabs}
\usepackage{multicol}
\usepackage{multirow}
\usepackage{verbatim}

\title{Beyond the Sentence: A Survey on Context-Aware Machine Translation with Large Language Models}

  \author{Ramakrishna Appicharla$^1$\thanks{\hspace{2.5px} Joint First Authors}, Baban Gain$^1$\footnotemark[1], Santanu Pal$^2$, Asif Ekbal$^3$ \\ $^1$Department of Computer Science and Engineering, Indian Institute of Technology Patna, India \\ $^2$Wipro AI, Lab45, London, UK \\ $^3$School of AI and Data Science, Indian Institute of Technology Jodhpur, India \\ {\tt \{ramakrishnaappicharla,gainbaban, santanu.pal.ju, asif.ekbal\} @gmail.com}}

\begin{document}
\maketitle
\begin{abstract}
Despite the popularity of the large language models (LLMs), their application to machine translation is relatively underexplored, especially in context-aware settings. This work presents a literature review of context-aware translation with LLMs. The existing works utilise prompting and fine-tuning approaches, with few focusing on automatic post-editing and creating translation agents for context-aware machine translation. We observed that the commercial LLMs (such as ChatGPT and Tower LLM) achieved better results than the open-source LLMs (such as Llama and Bloom LLMs), and prompt-based approaches serve as good baselines to assess the quality of translations. Finally, we present some interesting future directions to explore \footnote{TL; DR: Please refer to the Tables \ref{tab:prmompt-overview}, \ref{tab:ft-overview}, and \ref{tab:misc-overview} for the overview of the works covered in this survey.}.

\end{abstract}

\section{Introduction}
\label{sec:introduction}
Machine Translation (MT) translates a natural language sentence from one language into another without losing the meaning. It has seen tremendous improvements with the neural network-based methods \citep{sutskever2014sequence,bahdanau2014neural}, especially with transformer-based \citep{vaswani2017attention} models. However, most of the approaches model the translation at the sentence-level and often fail to translate the discourse phenomenon (such as pronouns and ellipses) \citep{voita-etal-2018-context,wang-etal-2023-survey} and produce inconsistent translations. Document-level (or context-aware) translation \citep{maruf-haffari-2018-document,zhang-etal-2018-improving,bawden-etal-2018-evaluating,agrawal-etal-2018-contextual,voita-etal-2019-good,huo-etal-2020-diving,li-etal-2020-multi-encoder,donato-etal-2021-diverse,maruf2021survey} addresses this issue by training a model to translate documents or paragraphs instead of sentences. There are two major approaches to building context-aware neural machine translation (NMT) models: concatenation-based \citep{tiedemann-scherrer-2017-neural,agrawal-etal-2018-contextual,junczys-dowmunt-2019-microsoft,zhang-etal-2020-long} approaches and multi-encoder \citep{zhang-etal-2018-improving,voita-etal-2018-context,kim-etal-2019-document,ma-etal-2020-simple} approaches. \citet{sun-etal-2022-rethinking} have shown that the standard transformer model has enough capacity to handle discourse-level phenomena effectively. However, one of the main limitations in building context-aware MT systems is the unavailability of document-level \citep{zhang-etal-2022-multilingual} parallel corpora.

\begin{figure}[t]
    \centering
    \includegraphics[width=1.0\linewidth]{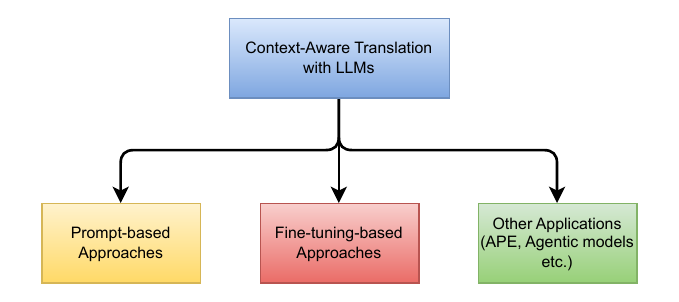}
    \caption{Overview of the survey presented in this work.}
    \label{fig:approaches}
\end{figure}

Large language models (LLMs) have become new paradigm to build various natural language processing (NLP) applications \citep{brown2020language,touvron2023llama2,zhang2022opt,JMLR:v24:22-1144,workshop2022bloom,alpaca} including machine translation (MT) \citep{zhang2023baylingbridgingcrosslingualalignment,gpt-neo,xu2024paradigmshiftmachinetranslation}. \citet{pang-etal-2025-salute} re-evaluated the six challenges \citep{koehn-knowles-2017-six} of the NMT models with LLMs and reported that data scarcity is not a significant challenge anymore because the LLMs are pre-trained on larger amount of data, especially for document-level translation. They fine-tuned Llama-2 \citep{touvron2023llama2} on the German-English language pair and reported that the LLMs fine-tuned with sentence-level data achieved good translation performance for longer sentences. Similarly, there have been several works to improve the document-level translation with LLMs. This work reports the current research regarding context-aware translation with LLMs. There have been surveys that focused either on non-LLM-based document-level translation \citep{maruf2021survey,peng2024handling} or on LLM-based sentence-level translation \citep{lyu-etal-2024-paradigm,gain2025bridginglinguisticdividesurvey}. The main aim of this survey is to provide a detailed review of the document-level MT with LLMs, which can help build more efficient models to handle the task of document-level translation.

The rest of the paper is organised as follows: Section \ref{sec:prompting} and Section \ref{sec:ft} describe the prompting-based and fine-tuning-based approaches (cf. Figure \ref{fig:approaches}), respectively. These two sections contain works exploring the context-aware translation with LLMs. In Section \ref{sec:other_applications}, we describe the works related to context-aware translation, such as APE, exploring the context usage, and developing an agentic framework for document-level translation. Finally, Section \ref{sec:conclusion} concludes the work with some potential future directions to explore.

\section{Prompting-based Approaches}
\label{sec:prompting}
The prompt is a natural language statement or instruction to steer an LLM's output in a specific direction. Prompting is an inference-time technique and does not require any training/fine-tuning of the LLM. Commonly used prompting-based approaches to \cite{brown2020language} build context-aware MT systems are (i). Zero-Shot Prompting (ZSP), and (ii). Few-Shot Prompting (or) In-Context Learning (ICL). In ZSP, the LLM is only given the prompt (such as \textit{Translate this sentence from English to German: <sentence>}) to carry out a specific task and no additional information is given on how to perform that task. ZSP might lead to suboptimal performance if the underlying LLM is not explicitly trained on the specific task. In ICL, the LLM is provided with a few examples (input-output pairs) and the prompt. The examples provide necessary context to achieve the task and improve performance effectively. ZSP and ICL-based approaches are followed to develop context-aware MT systems using LLMs. This section mainly describes the works that follow different prompting-based approaches (cf. Table \ref{tab:prmompt-overview}) to build and evaluate context-aware MT systems.

\begin{table*}[!ht]
    \centering
     \resizebox{1.0\linewidth}{!}{
    \begin{tabular}{ccc}
        \toprule
        \textbf{Work} & \textbf{LLMs} & \textbf{Language Pairs} \\
        \midrule
        \citet{wang-etal-2023-document-level} & GPT-3.5, GPT-4 & \begin{tabular}[c]{@{}c@{}} Zh $\rightarrow$ En, En $\rightarrow$ De, En $\rightarrow$ Ru \end{tabular} \\
        \cmidrule{1-3}
        \citet{hendy2023good} &  \begin{tabular}[c]{@{}c@{}}ChatGPT,\\ GPT-3.5 (text-davinci-002, 003) \end{tabular} & En $\leftrightarrow$ De \\
        \cmidrule{1-3}
        \citet{moslem-etal-2023-domain} &  \begin{tabular}[c]{@{}c@{}}ChatGPT,\\ GPT-3.5 (text-davinci-003),\\ Bloom, Bloomz \end{tabular} & En $\rightarrow$ \{Ar, Zh, Fr, Rw, Es\} \\
        \cmidrule{1-3}
        \citet{zhang2023prompting}  & GLM & Zh $\rightarrow$ En \\
        \cmidrule{1-3}
        \citet{wu-hu-2023-exploring}  & GPT-3.5 & En $\leftrightarrow$ Zh \\
        \cmidrule{1-3}
        \citet{bawden-yvon-2023-investigating}  & Bloom, Bloomz & En $\leftrightarrow$ Fr \\
        \cmidrule{1-3}
        \citet{karpinska-iyyer-2023-large}  & GPT-3.5 (text-davinci-003) & \begin{tabular}[c]{@{}c@{}} \{En, De, Fr, Ru, Cs, Ja\} $\rightarrow$ Pl;\\ \{En, De, Fr, Ru, Pl, Zh\} $\rightarrow$ Ja;\\ \{De, Fr, Ru, Pl, Ja, Zh\} $\rightarrow$ En \end{tabular} \\
        \cmidrule{1-3}
        \citet{sia-duh-2023-context}  & GPT-Neo, Bloom, XGLM & En $\rightarrow$ \{Fr, De, Pt\} \\
        \cmidrule{1-3}
        \citet{cui-etal-2024-efficiently}  & Qwen, Baichuan & En $\leftrightarrow$ De, En $\leftrightarrow$ Zh \\
        \cmidrule{1-3}
        \citet{enis2024llm} & Claude 3 Opus & Yo $\rightarrow$ En \\
        \cmidrule{1-3}
        \citet{hu2025source} & \begin{tabular}[c]{@{}c@{}} Qwen2.5, Llama-3.1,\\ GPT-4o-mini \end{tabular} & \begin{tabular}[c]{@{}c@{}} En $\rightarrow$ \{Zh, Cs, De, Hi, Is, Ja, Ru, Es, Uk\};\\ Cs $\rightarrow$ Uk;\\ Ja $\rightarrow$ Zh \end{tabular} \\
        \bottomrule
    \end{tabular}
    }
    \caption{List of prompt-based approaches covered in this work.}
    \label{tab:prmompt-overview}
\end{table*}

\citet{wang-etal-2023-document-level} conducted experiments to evaluate the discourse modelling abilities of LLMs through ZSP. Specifically, they designed three prompts where two prompts are designed to translate a document in a single conversation turn \footnote{Both the prompts are similar, but one of the prompts includes the tags to indicate the sentence boundaries.} and one prompt to translate the document sentence-by-sentence in multiple conversation turns. All the experiments are conducted on ChatGPT \footnote{Experiments are conducted on both GPT-3.5 and GPT-4 versions.}. They conducted experiments on the online chat version of ChatGPT. They reported that all three prompts performed equally well, and even the sentence-by-sentence translation prompt results were promising because ChatGPT is generally good at remembering the context within a single chat session. They also report that ChatGPT (GPT-4) is better in discourse-level translation than the commercial MT systems, such as Google Translate, DeepL Translator, and Tencent TranSmart.

\citet{hendy2023good} conducted experiments in ZSP and ICL settings on GPT-3.5 (text-davinci-003) to study how the context affects the translation performance. Specifically, in the ZSP setting, each document is translated sentence-by-sentence, with the number of context length varying (in the range of 1-32, where one is only one sentence is translated at a time without any context, and 32 is all 32 sentences are translated at the same time in a sentence-by-sentence fashion). They observed that increasing the length of the context window leads to better performance. They also conducted ICL with five examples and a context window size of 10 sentences. They report that the few-shot learning is redundant for document-level translation, as the source document provides sufficient context for effective translation.

\citet{moslem-etal-2023-domain} conducted experiments on document-level translation with fuzzy matches (also known as translation memories) as examples for ICL. The experiments are performed on GPT-3.5 (text-davinci-003) and Bloom \citep{workshop2022bloom} and Bloomz \citep{muennighoff-etal-2023-crosslingual} LLMs, and documents are translated sentence-by-sentence. They reported that ICL (up to 5 examples) performs better than ZSP and examples selected randomly.

\citet{zhang2023prompting} studied whether the knowledge can be transferred from examples selected from a sentence-level corpus to translate document-level data. They conducted experiments on GLM-130B \citep{zeng2022glm} LLM in a 1-shot prompting setting with examples taken from the sentence-level corpus and used to translate a document chunk (containing about four sentences). The results show that the performance of 1-shot prompting is better than ZSP in both document-BLEU and document-specific \citep{sun-etal-2022-rethinking} metrics.

\citet{wu-hu-2023-exploring} conducted experiments on GPT-3.5 to translate documents sentence-by-sentence using ZSP. They reported that translating sentences as multi-turn dialogue yields better results than translating sentences in isolation. They also prompted the model to translate and revise the translations in a single turn. However, this approach achieves the lowest performance because the LLM cannot revise the results, as there is no clear or specific instruction to guide the correction process.

\citet{bawden-yvon-2023-investigating} conducted experiments on Bloom \citep{workshop2022bloom} and Bloomz \citep{muennighoff-etal-2023-crosslingual} LLMs in 1-shot setting. They selected two types of examples, random and previous dialogue utterances. They report that the previous dialogue example gives the best score; the difference between results from random example selection and the previous dialogue selection is very low. It is unclear if the context of the previous sentence is better than that of the random context. These results are in line with the previous findings on non-LLM DocMT \citep{li-etal-2020-multi-encoder,sun-etal-2022-rethinking,appicharla-etal-2023-case,appicharla-etal-2024-case} models.

The availability of document-level parallel corpora is less than that of sentence-level parallel corpora, and even less so in literary translation. \citet{karpinska-iyyer-2023-large} created paragraph-level parallel corpora for 18 language pairs from the publicly available novels and short stories (both the source and translations are available for some novels and short stories. None of the collected data is synthetic). They performed ICL with five examples on GPT-3.5 (text-davinci-003) in three different settings: translating sentence-by-sentence, paragraph-by-paragraph, and paragraph-by-paragraph with sentence markers to distinguish between the sentences. They also performed MQM \citep{freitag2021experts} and annotated the translations with seven different types of error categories\footnote{1. Mistranslation, 2. Grammar, 3. Untranslated words, 4. Inconsistency, 5. Register (formal and informal use of the language), 6. Format (correct usage of punctuation), 7. Additions and omissions} along with the human evaluation. They reported that humans prefer the paragraph-to-paragraph setting over other outputs. However, the common mistakes observed are mistranslations and grammatical errors.

\citet{sia-duh-2023-context} studied how the examples chosen for ICL affect the coherence of the translations. To examine how the context (provided in examples) affects the document-level translation, they compared five randomly selected examples with five previous sentences (moving window approach). Source sentences and their gold translations are provided as examples, and the document is translated sentence-by-sentence. All experiments are conducted on GPT-Neo \citep{gpt-neo}, Bloom \citep{workshop2022bloom}, XGLM \citep{lin2021few}, and they reported that the previous 5-sentence context achieves the best results compared to the random selection. They also observed that selecting examples randomly from the same document performs better than selecting examples randomly from outside the document. However, the order of these examples (such as keeping the same order as in the document vs shuffling the order of these sentences) does not affect the performance. Finally, they also noted that keeping the static context (such as selecting the first five sentences and using them throughout the translation of the document) performs worse than random sampling. This shows that choosing the correct context is essential for efficient translations, but the order of these examples can vary.

\citet{cui-etal-2024-efficiently} proposed a context-aware prompting method to address the issue of incoherent translations generated by the LLMs via an ICL approach. Specifically, they proposed a Context-Aware Prompting method (CAP) to select the examples for ICL. The proposed CAP method consists of three steps. In the first step, sentences similar to the current sentence are extracted based on the sentence-level attention scores (dynamic context window). In the second step, the extracted context from the first step is summarised and is used to retrieve similar sentences from an external database. Finally, the retrieved sentences are used as examples for the ICL. The experiments are conducted on Baichuan \citep{yang2023baichuan} and Qwen \citep{bai2023qwen} LLMs. They also compared the proposed approach with different prompting strategies, such as ZSP and ICL, with random examples and the previous three sentences. The proposed approach achieves the best results. However, the performance of the ZSP is also very similar to the proposed approach. The larger LLMs (in the range of 14-72B parameters) are more robust to the random context than the smaller LLMs in the case of random context, where the examples are selected randomly from the database. When models are evaluated on the zero pronoun translation (ZPT) task, the ZSP perform poorly compared to other ICL approaches. This shows that the context is essential for effective context-aware translation.

\citet{enis2024llm} conducted experiments on Claude LLM for a low-resource Yoruba-English language pair. Specifically, they follow ICL to translate the Yoruba news documents (crawled from the BBC) into English sentence-by-sentence. The example input to the LLM consists of a document split into sentences and their corresponding translation, and the LLM is prompted to translate another document based on the given example sentence-by-sentence. They report that the Claude LLM performs significantly better in automatic metrics (i.e., BLEU and ChrF++) than traditional MT systems, such as Google Translate and NLLB \citep{costa2022no} models. However, they did not perform an in-depth context-aware analysis of the translations that were obtained.

\citet{hu2025source} conducted experiments on translating a document in a multi-turn conversation setting. Specifically, they compared the multi-turn translation performance to single-turn (where the whole document is translated in a single turn) and segment-level translation (where the document is split into multiple paragraphs and translated independently). They observed that single-turn translation results are subpar due to the omission errors when translating long documents, compared to other approaches. The multi-turn translation setting achieves the best results due to translations from previous turns being cached and used as context. They also fed the whole source document before the translation process to further enhance the translation quality of the multi-turn translation setting. All the experiments were conducted on GPT-4o-mini \citep{hurst2024gpt}, Llama-3 \citep{grattafiori2024llama}, and Qwen-2.5 \citep{yang2024qwen2} LLMs. Finally, they also observed that the performance could improve with ICL (they used three examples in their experiments).

\section{Fine-tuning-based Approaches}
\label{sec:ft}

\begin{table*}[!ht]
    \centering
     \resizebox{1.0\linewidth}{!}{
    \begin{tabular}{ccc}
        \toprule
        \textbf{Work} & \textbf{LLMs} & \textbf{Language Pairs} \\
        \midrule
        \citet{zhang-etal-2023-machine} & GPT-Neo, OPT, Llama-2, Bloomz & Fr $\rightarrow$ En \\
        \cmidrule{1-3}
        \citet{wu2024adapting} & \begin{tabular}[c]{@{}c@{}} GPT-3.5-Turbo, GPT-4-Turbo,\\ Llama-2, Vicuna, Bloom \end{tabular} & En $\leftrightarrow$ \{Ar, De, Fr, It, Ja, Ko, Nl, Ro, Zh\} \\
        \cmidrule{1-3}
        \citet{li2024enhancing} & Llama-2 & En $\leftrightarrow$ \{Fr, De, Es, Ru, Zh\} \\
        \cmidrule{1-3}
        \citet{lyu-etal-2024-dempt} & Llama-2, Bloomz & \{Fr, De, Es, Ru, Zh\} $\rightarrow$ En \\
        \cmidrule{1-3}
        \citet{wang-etal-2024-benchmarking} & GPT-3.5, GPT-4, Llama-2 & Zh $\rightarrow$ En; En $\rightarrow$ \{De, Ru\} \\
        \cmidrule{1-3}
        \citet{kudo-etal-2024-document} & Llama-2, Mistral & En $\rightarrow$ Ja, Ja $\rightarrow$ Zh \\
        \cmidrule{1-3}
        \citet{sung-etal-2024-context} & GPT-4o, Gemma & En $\leftrightarrow$ Ko \\
        \cmidrule{1-3}
        \citet{yang-etal-2024-optimising} & Llama-3 & En $\leftrightarrow$ \{De, Nl, Pt-Br\} \\
        \cmidrule{1-3}
        \citet{pombal-etal-2024-improving} & Tower-Chat & En $\leftrightarrow$ \{Fr, De, Nl, Ko, Pt-Br\} \\
        \cmidrule{1-3}
        \citet{wu-etal-2024-choose} & Llama-2 & En $\rightarrow$ Zh \\
        \cmidrule{1-3}
        \citet{elshin-etal-2024-general} & YandexGPT & En $\rightarrow$ Ru \\
        \cmidrule{1-3}
        \citet{zafar-etal-2024-setu-adapt} & Llama-3 & \{Fr, De\} $\rightarrow$ En \\
        \cmidrule{1-3}
        \citet{luo-etal-2024-context} & Chinese-Llama-2 & Zh $\rightarrow$ En \\
        \bottomrule
    \end{tabular}
    }
    \caption{List of fine-tuning-based approaches covered in this work.}
    \label{tab:ft-overview}
\end{table*}

The ability of LLMs to accurately translate context-aware data can be further improved over the prompt-based approaches via fine-tuning (FT). The LLM can be further fine-tuned based on data from a specific language pair or domain data. Generally, fine-tuning the LLM for a particular downstream task has shown improvements \citep{brown2020language} but at the expense of more computational requirements. There have been several works on fine-tuning LLM to improve context-aware translation (cf. Table \ref{tab:ft-overview}) performance.

\citet{zhang-etal-2023-machine} is one of the earlier works that explored the translation performance of different LLMs with prompting and fine-tuning-based approaches for both sentence and document-level settings. They experimented on GPT-Neo \citep{gpt-neo}, OPT \citep{zhang2022opt}, Llama-2 \citep{touvron2023llama2}, XGLM \citep{lin2021few}, and Bloomz \citep{muennighoff-etal-2023-crosslingual} LLMs. They studied ZSP and ICL settings and fine-tuned the LLMs with QLoRA \citep{dettmers2023qlora}. They conducted experiments on the French-to-English language pair. They observed that fine-tuning consistently achieves better results than prompt-based approaches in sentence and document-level translation settings. Further, they reported that fine-tuning the LLMs with a document length of 10 sentences achieves better results than other models trained with 5 or 15 sentences per document. They noted that fine-tuning multilingual LLMs (XGLM and Bloomz) achieves significantly better results than prompting.

\citet{wu2024adapting} compared various non-LLM-based models \citep{costa2022no,sun-etal-2022-rethinking,wu-etal-2024-importance}, close-source models (such as Google Translate, GPT-3.5-Turbo, and GPT-4-Turbo) with open-source LLMs, such as Llama-2 \citep{touvron2023llama2}, Bloom \citep{workshop2022bloom}, and Vicuna \citep{vicuna2023} that are fine-tuned with document-level parallel corpora. They fine-tune the LLMs in a two-step process where the LLM is trained on the monolingual documents \citep{xu2023paradigm} followed by parallel documents. They used the previous three sentences as the context. They followed both full fine-tuning (FFT) and Low-rank adaptation (LoRA) \citep{hu2022lora} to fine-tune the LLMs. They trained the models on nine language pairs. They reported that LoRA performs better than FFT in bilingual scenarios (where each language pair is trained separately) and FFT performs better in multilingual scenarios. They also observe that FFT requires 1\% of data and LORA requires 10\% to achieve identical performance. However, GPT-4-Turbo achieves the best results across all language pairs with NLLB, and fine-tuned Bloom-7B models achieve the next best results for English-to-other and other-to-English language pairs, respectively. They also experimented with two different inference approaches: (i). using previously translated sentences as the context, and (ii). translating each sentence in the context independently. They report that using the previously translated context leads to poor translation quality due to error propagation, but translating each sentence achieves good results, but is computationally more expensive.

LLMs fine-tuned on sentence-level corpora cannot translate the larger documents (containing more than 512 tokens). To effectively translate longer documents, \citet{li2024enhancing} proposed combining sentence and document-level instructions of varying length to fine-tune LLMs. The documents are split into sub-documents of length $[512, 1024, 1536, 2048]$ tokens, and they are combined with sentence-level instructions to create translation-mixed instructions. They followed the Alpaca instruction format \citep{alpaca} to form these translation instructions and fine-tuned the Llama-2 \citep{touvron2023llama2}. The model fine-tuned with translation-mixed instructions (sub-document length $1024$ setting) achieves the best results compared to other fine-tuning settings when translating from different languages to English. However, when translating from English to other languages, the non-LLM-based DocNMT model \citep{li2023p} achieves the best results over the LLM-based models. The proposed approach also improves the discourse-level translation accuracy with longer length instructions (specifically both $1024$ and $1536$ settings), achieving the best results on discourse phenomenon evaluation.

During the fine-tuning of the LLM for context-aware translation, the input consists of concatenated context and source sentences. However, the context and the source sentences are given the same priority when concatenated. This might not be optimal for effective translation because the source should be given more priority over the context. To address this issue, \citet{lyu-etal-2024-dempt} proposed a technique (named Decoding-enhanced Multi Prompt Tuning) which enables the LLMs to discriminate the context and the source sentences to enhance the translation performance. Specifically, there are three steps for training the model. In the first step, all the context sentences are concatenated to form a single sequence to be encoded. In the second step, the current source sentence is encoded based on the activations obtained from the previous step. The resulting activations from the first and second steps are used to predict the target sequence probabilities in the third step. The decoding is further enhanced by combining the activations from the above three steps. Trainable prompts are incorporated during each phase, and deep prompt-tuning \citep{li2021prefix,liu2021p} is followed to train the models. They conducted experiments on Llama-2 \citep{touvron2023llama2} and Bloomz \citep{muennighoff-etal-2023-crosslingual} LLMs and context window set to a maximum of 256 tokens. The proposed approach significantly outperforms the non-LLM-based models \citep{bao-etal-2021-g,sun-etal-2022-rethinking} and the LLM-based models trained with concatenated context (i.e., without multi-phase training) and without context (i.e., fine-tuning without the context). They report that the proposed approach is more robust to context changes and can perform well on longer context inputs.

\citet{wang-etal-2024-benchmarking} constructed instruction-based datasets for fine-tuning LLMs for context-aware translation. Specifically, they followed Alpaca \citep{alpaca} format to create instructions and added them to the existing document-level corpora to convert them into instruction-based corpora. They have used GPT-4 to generate instructions (eg, Translate the following sentence from English to German: \textit{source sentence}) and created datasets for multiple languages (Chinese-English, Russian-English, and German-English) and domains (News, Subtitles, TED talks, Parliamentary proceedings, Novels). The sentences are concatenated up to the maximum supported length of the LLM during the inference. They compared the results from different commercial MT systems (Google Translate, DeepL, and Tencent Translate), closed-source LLMs (GPT-3.5, GPT-4), and open-source LLMs (Llama-2). Results show that GPT-4 (with 8K input length) achieves the best results (except in the Novels domain) over the other systems. They also fine-tuned Llama \citep{touvron2023llama} on the prepared corpora to study the effect of the document lengths. They observed that the model trained with a maximum length of 2048 tokens achieves the best results over the model trained with a maximum length of 4096 tokens and over the sentence-level model. They report that varying the document length during the fine-tuning achieves better results over the models trained on fixed-length documents. Finally, they experimented with multi-task learning where the auxiliary task is text completion (i.e., generating the following sentence based on given context). They reported that multi-task learning improves translation performance, especially in domains such as literature.

\citet{kudo-etal-2024-document} conducted experiments on Minimum Bayes Risk (MBR) decoding \citep{eikema-aziz-2020-map} and LLM-based reranking to improve the translation performance. They initially trained non-LLM-based and LLM-based MT models for sentence-level translation. During the decoding, the best hypotheses are selected using MBR decoding with COMET-22 \citep{rei-etal-2022-comet} as the utility function. After extracting the 30 best hypotheses from the MBR decoding step, LLM (fine-tuned on document-level corpora) selects the hypotheses with the highest likelihood scores. They used Llama-2 \citep{touvron2023llama2} and Mistral-7B \citep{jiang2023mistral7b} LLMs for translation and Mistral-7B for reranking. They report that MBR-enhanced decoding is more effective than LLM-based reranking.

\citet{sung-etal-2024-context} proposed using conversation summarisation to manage the context length effectively. Specifically, they used the GPT-4o mini model to generate the conversation summary. The generated summary and the two recent source-target pairs are used as the context to fine-tune the Gemma \citep{gemmateam2024gemma2improvingopen} LLM with LoRA \citep{hu2022lora}.

\citet{yang-etal-2024-optimising} used a sliding window approach to improve the performance of chat translation. The first sentence will be translated without the context and stored as context for further translations. The subsequent sentences are translated based on the cached source sentences. If the cache window size exceeds the limit, the earliest source sentence is removed, and the current input is stored. Llama-3 \citep{grattafiori2024llama} is fine-tuned based on the sliding window approach. They experimented on six language pairs with three different window sizes (1-3) and reported that window size 2 is the most optimal. They also noted that fine-tuning with multilingual data gives better results than bilingual data.

\citet{pombal-etal-2024-improving} experimented on fine-tuning and quality aware decoding \citep{fernandes-etal-2022-quality,freitag-etal-2022-high} strategies to improve the chat translation performance. The model is fine-tuned on the chat data during the fine-tuning, and all the previous source-target pairs are used as context. For the quality-aware decoding, they conducted (MBR) decoding \citep{eikema-aziz-2020-map} with COMET \citep{rei-etal-2022-comet} and Context-COMET \citep{agrawal-etal-2024-assessing} as the utility functions on a sample of 100 candidate translations. They used Tower LLM \citep{alves2024toweropenmultilinguallarge} and reported that MBR decoding combined with fine-tuning achieves better results than ZSP or ICL settings.

\citet{wu-etal-2024-choose} used continual pre-training, supervised fine-tuning, and contrastive preference optimisation (CPO) \citep{xu2024contrastivepreferenceoptimizationpushing} to improve the long text translation performance of LLMs. They conducted experiments on Llama-2 \citep{touvron2023llama2} and observed that CPO improves the translation when combined with supervised fine-tuning. They also reported that MBR decoding with COMET achieves the best results.

\citet{elshin-etal-2024-general} explored deep prompt-tuning \citep{li2021prefix,liu2021p}, and CPO \citep{xu2024contrastivepreferenceoptimizationpushing} for paragraph-level translation for video subtitles data. They followed the curriculum learning where the LLM is first tuned with sentence-level data, then with paragraph-level examples towards the end of training. They reported that CPO with a curriculum learning based model achieves the best results.

\citet{zafar-etal-2024-setu-adapt} used sentence-transformers \citep{reimers-gurevych-2019-sentence} to extract 3 examples and used them to fine-tune Llama-3 \citep{grattafiori2024llama} for chat translation. However, the results of non-LLM-based models (such as NLLB \citep{costa2022no}) achieved the best results over the fine-tuned Llama-3.

\citet{luo-etal-2024-context} fine-tuned Chinese-Llama-2 \citep{cui2024efficienteffectivetextencoding} for Chinese-English literary translation. Their framework for fine-tuning consists of continual pre-training with monolingual data, concatenated source-target sentence pairs, and supervised fine-tuning with context-aware corpus \citep{guo-etal-2024-novel}. During the inference, all the previously translated sentences and sentences similar to the source are used to maintain consistency and style-related information.

\section{Other Applications}
\label{sec:other_applications}
We describe other miscellaneous works related to context-aware translation, such as post-editing for context-aware translation, using LLMs to evaluate the performance, etc. Table \ref{tab:misc-overview} shows the overview of the miscellaneous works.

\begin{table*}[!ht]
    \centering
    \resizebox{1.0\linewidth}{!}{
    \begin{tabular}{ccc}
        \toprule
        \textbf{Work} & \textbf{LLMs} & \textbf{Language Pairs} \\
        \midrule
        \midrule
        \multicolumn{3}{c}{\textbf{Automatic Post-Editing}} \\
        \midrule
        \midrule
        \citet{koneru-etal-2024-contextual} & Llama-2 & En $\rightarrow$ De \\
        \cmidrule{1-3}
        \citet{li-etal-2025-enhancing-large} & Llama-3 & En $\rightarrow$ De \\
        \cmidrule{1-3}
        \citet{dong2025intermediatetranslationsbetterone} & Llama-3, Mistral-Nemo-Instruct & En $\leftrightarrow$ \{Fr, De, Ru, Es, Zh\} \\
        \midrule
        \midrule
        \multicolumn{3}{c}{\textbf{Agentic Framework}} \\
        \midrule
        \midrule
        \citet{wang2024delta} & GPT-3.5-Turbo, GPT-4o-mini, Qwen2 & En $\leftrightarrow$ \{Fr, De, Zh, Ja\} \\
        \cmidrule{1-3}
        \citet{briakou-etal-2024-translating} & Gemini-1.5-Pro & En $\rightarrow$ \{De, Cs, Ru, Ja, He, Uk, Zh\} \\
        \cmidrule{1-3}
        \citet{wu2024perhaps} & GPT-4 & Zh $\rightarrow$ En \\
        \cmidrule{1-3}
        \citet{guo2025doc} & Qwen2.5, Llama-3.1 & En $\leftrightarrow$ \{Fr, De, Zh, Ja\} \\
        \midrule
        \midrule
        \multicolumn{3}{c}{\textbf{Miscellaneous Works}} \\
        \midrule
        \midrule
        \begin{tabular}[c]{@{}c@{}} \citet{petrick-etal-2023-document}\\ (Combining sentence-level NMT\\ with document-level LM) \end{tabular} & Llama & En $\rightarrow$ \{De, It\} \\
        \cmidrule{1-3}
        \begin{tabular}[c]{@{}c@{}} \citet{sun-etal-2025-fine}\\ (LLM-as-Judge) \end{tabular} & GPT-4, Vicuna, Mistral & En $\leftrightarrow$ \{De, Zh\} \\
        \cmidrule{1-3}
        \begin{tabular}[c]{@{}c@{}} \citet{mohammed2024analyzing}\\ (Analysis of context usage) \end{tabular} & \begin{tabular}[c]{@{}c@{}} EuroLLM, Llama-2,  ALMA,\\ TowerBase, TowerInstruct \end{tabular} & En $\rightarrow$ \{Fr, De\} \\
        \bottomrule
    \end{tabular}
    }
    \caption{List of APE, Agentic framework, and other miscellaneous approaches covered in this work.}
    \label{tab:misc-overview}
\end{table*}

\citet{petrick-etal-2023-document} proposed combining a sentence-level MT system with a document-level LM, which is only trained with the document-level monolingual data, to improve the context-aware translation performance. There are three ways to estimate the target token probability during the decoding phase: (i). From the sentence-level MT model, (ii). From the monolingual document-level LM model, and (iii). The internal LM that is learned by the MT model \citep{herold-etal-2023-improving}. The probability of a target token is effectively estimated by combining these three probabilities. They primarily conducted experiments on a small LM (35M parameters, trained on NewsCrawl corpus \footnote{\url{https://data.statmt.org/news-crawl/}}) and reported that the proposed approach generally yields better results without much computational overhead. They also noted that using Llama \citep{touvron2023llama} (13B parameters) instead of the small LM perform better.

The existing metrics (such as BLEU, ChrF) are ill-suited \citep{muller-etal-2018-large,post2023escaping} for evaluating the context-aware translation. \citet{sun-etal-2025-fine} followed `LLM-as-a-judge' \citep{zheng2023judging} paradigm to asses the translation quality of context-aware translation. They argue that the ideal context-aware evaluation metric should be (i). Context-aware (to capture the document-level coherency and accuracy), (ii). Structured (to evaluate fluency, accuracy, and coherence separately), and (iii). Interpretable (easy to understand and identify the errors clearly). To this end, they proposed to evaluate the translations based on four metrics: (i). Fluency (rating translations from 1-5), (ii). Content errors (mistranslations, omissions, and additions), (iii). Lexical cohesion errors (incorrect vocabulary usage, missing synonyms, and overuse of words), and (iv). Grammatical cohesion errors (pronouns, conjunctions, sentence-linking structure mistakes). They used GPT-4 as the judge to evaluate the translations based on the above four metrics. They generated translations using instruction-tuned LLMs such as Vicuna \citep{vicuna2023} and Mistral-7B \citep{jiang2023mistral7b}. They generated the translations in two settings, concatenating previous k (in the range of 1-3) sentences, translating the resulting sequence, and translating the entire document in a single turn. They reported that concatenated sentence translation achieves best results (k=3) over translating the entire document in a single turn in terms of BLEU \citep{liu-etal-2020-multilingual-denoising} score. However, translating the entire document in a single turn achieves the best results for the proposed context-aware metrics. This shows that the existing automatic metrics are ill-suited for evaluating context-aware translations.

\citet{mohammed2024analyzing} investigated how sensitive the LLMs are towards the correct context and how well they utilise the context during the context-aware translation. They followed two approaches to analyse the context utilisation: (i). Perturbation analysis: Studying the model's robustness by examining the translation quality and pronoun resolution performance on different types of context. (ii). Attribution analysis: Analysing the contribution of relevant parts of the context during the translation through ALTI-Logit (calculates the contribution of each source token on a particular target token) \citep{ferrando2023explaining} and Input-Erasure (measures the change in model's prediction when removing parts of the input) \citep{li2016understanding} attribution methods. They conducted experiments on nine different LLMs on English-to-French and English-to-German language pairs. In the perturbation analysis, they experimented on three different types of context: (i). Gold (previous source-target pairs as the context), (ii). Perturbed (randomly samples sentence pairs from other documents), (iii). Random (uniform sampling of random tokens from the model's vocabulary) and utilised contrastive testsets \citep{muller-etal-2018-large,lopes-etal-2020-document,post2023escaping} for pronoun resolution experiments. They followed ICL with five previous source-target pairs as context. All the models are robust to random context (with respect to automatic metrics), which shows the lack of proper context utilisation. Still, the effects of random and perturbed contexts are more observed in the contrastive test sets. They report a need for explicit context-aware fine-tuning to utilise the context better.

\citet{wang2024delta} proposed a document-level translation agent with four memory components. These memory components are: (i). Proper noun records: This contains previously translated source-target noun pairs. The proper noun records are updated when a new proper noun is identified, and if the noun already exists in memory, then its corresponding translation will be used. This is to maintain the consistency of translation across the whole document. (ii). Bilingual summary: This contains both the source and target summaries. The summary components ensure coherency and help produce more consistent translations. (iii). Long-term memory (iv). Short-term memory: Both the long-term and short-term memories contain the previously translated source-target sentence pairs. Specifically, the short-term memory is designed to capture the context of immediate sentences surrounding the current sentence, and the long-term memory is designed to capture the broader context. The information from these four memory components is integrated into a prompt for translating the document sentence-by-sentence, ensuring that all the sentences in the source document are translated while maintaining consistency and fluency. They conducted experiments on GPT-3.5, GPT-4 \citep{brown2020language,achiam2023gpt} and Qwen \citep{bai2023qwen} LLMs.

\citet{briakou-etal-2024-translating} proposed a chain-of-thought \citep{wei2022chain} approach to document translation. The framework consists of four phases, during which the source document is translated and refined to improve the translation's accuracy and fluency. They conducted experiments on Gemini 1.5 Pro \citep{team2024gemini} LLM, and the proposed approach improves the translation performance over the ZSP approach.

\citet{wu2024perhaps} proposed a multi-agent virtual agentic framework for literary translation. Their approach emulates various hierarchies in the translation workbench (eg, translators, proofreaders, editors, etc.). Their framework can effectively translate very long literary texts and is preferred in human evaluation metrics, but fails to translate shorter texts.

One of the main issues with translating an entire document (Doc-to-Doc translation) with LLMs is that some of the sentences in the source document are not being translated. To translate every source sentence in the given document, sentences need to be processed sentence-by-sentence, often leading to consistency and fluency issues (such as missing context coherence). To ensure the translation of every sentence without the problems of consistency, \citet{guo2025doc} introduced a prompt-based agent that uses an incremental sentence-level forced decoding strategy. Specifically, the document is translated sentence-by-sentence to ensure all the source sentences are translated. However, the input to the LLM consists of two source sentences (i.e., $s_{(i-1)}$ and $s_i$) with the output of the previous source sentence (i.e., $t_{(i-1)}$) acting as a mandatory decoding segment. This forces the LLM to regenerate the previously translated sentence and maintain fluency while translating the entire document. Finally, the translation of the current source sentence (i.e., $t_i$) is concatenated to form the final translation. They also augmented the summary of both source and target documents in the form of memory to enhance discourse translation. They conducted experiments on Qwen \citep{bai2023qwen} and Llama-3 \citep{grattafiori2024llama} LLMs. They observed no significant difference in performance when more than one previous sentence was forced decoded.

Automatic post-editing (APE) \citep{pal-etal-2016-neural} is an effective way to improve the translation performance of any MT system. Some works have explored the APE for document-level translation with LLMs. \citet{koneru-etal-2024-contextual} proposed to use LLMs to post-edit the translations from NMT models. The LLM is fine-tuned on the (source, hypothesis, reference) triplet, where the hypothesis is generated from an NMT model. They trained the APE models for both sentence and document-level inputs. Specifically, the document-level sentences are translated sentence-by-sentence independently and concatenated to give as input to the APE model. They used Llama-2 \citep{touvron2023llama2} and followed LoRA \citep{hu2022lora} to train the APE models. They report that APE is necessary and significantly improves the translation performance over both the ICL approach and fine-tuned LLMs for the translation in terms of automatic and context-aware evaluation metrics. The performance of APE can be further improved by using gold context rather than using the model-generated outputs for document-level MT.

Similar to \citet{koneru-etal-2024-contextual}, \citet{li-etal-2025-enhancing-large} also employed LLM to post-edit the translation from non-LLM-based sentence-level MT models. They created the synthetic data to train the LLM by translating monolingual document-level target data with a target-to-source NMT model, and then using the created synthetic source, they further created the synthetic target data using a source-to-target NMT model. This process aims to generate synthetic-natural target data, which is used as supervised training data to train LLM. They fine-tuned the LLM with two objectives, one using only the synthetic and clean side data and another using synthetic target-source and clean target data. Although using source and target data to train the APE model is intuitive, they observed that it leads to poor performance. The reason for this is that the APE model is trained with synthetic source data, but during the inference, the clean source data is fed to the model along with the output (synthetic target) of the NMT model. They further proposed two approaches to address this issue: (i). The cascade approach is where the LLM is fine-tuned to perform APE for synthetic source data, and the resulting output is used along with the synthetic target data to perform APE for the synthetic target data (ii). A continuous pre-training approach is where the LLM is initially pre-trained with synthetic and natural source and target data and then fine-tuned with the synthetic source-target and clean target data. They conducted experiments on Llama-3 \citep{grattafiori2024llama} and reported that the proposed APE models outperform the Llama models that are only trained for translation in terms of both automatic and context-aware evaluation metrics. Interestingly, there is only a minute difference in the performance between the APE models trained only with the target data compared to those trained with both source-target data (i.e., cascade and continual pre-training approaches). Similarly, \citet{thai-etal-2022-exploring} fine-tuned GPT-3 to post-edit the MT outputs generated by Google Translate.

\citet{dong2025intermediatetranslationsbetterone} fine-tuned Llama-3 and Mistral-Nemo-Instruct \footnote{\url{https://mistral.ai/news/mistral-nemo}} to post-edit the document-level translations. Initially, they generated two sets (sentence-level and document-level) of translations and used them to train an APE model. They observed that the sentence-level translations reduce the hallucinations and the document-level translations reduce inconsistencies. Hence, using the translations from both these settings improves the refinement process of the document-level translations.

\section{Conclusion and Future Directions}
\label{sec:conclusion}
We have presented a literature review of the prior works that explored context-aware machine translation with LLMs. Prompt-based approaches (zero-shot and few-shot prompting) have shown that the LLMs can effectively handle context-aware translation tasks, and fine-tuning with the available corpora yields state-of-the-art performance for many language pairs (such as English, French, German, Spanish, and Chinese). The performance of closed-source LLMs (such as ChatGPT and Tower LLM) is better than that of open-source LLMs (such as Llama and Bloom) in prompt-based approaches. However, fine-tuning Llama or Bloom LLMs achieved nearly the same performance as ChatGPT or Tower LLM. LLMs are also employed in an automatic post-editing task to post-edit the translations and have shown good performance in capturing the discourse-level phenomenon. We observed that context-aware translation with LLMs is a relatively new research direction, and little work has been done. To this end, we suggest the following future research directions:

\paragraph{Context-aware translation for low-resourced languages:} Unlike the sentence-level corpora, document-level corpora are not available for many language pairs, but large amounts of monolingual document-level corpora are available. LLMs can be used to build context-aware translation systems for language pairs with little to no document-level corpora through prompting or automatic post-editing.

\paragraph{Translation agents:} Agentic framework \citet{wu2024perhaps,briakou-etal-2024-translating} is another interesting research direction to build context-aware MT systems. Different aspects of translation can be handled by various agents (such as lexical translation consistency, overall fluency, keeping track of context-aware words and phrases, etc.), similar to \citet{wang2024delta}.

\paragraph{Context-aware evaluation:} Robust and interpretable evaluation metrics are essential to assess the quality of machine translation outputs. LLMs can significantly affect \citep{agrawal-etal-2024-assessing} the evaluation of MT systems. Future directions can explore using LLMs with minimal supervised data to effectively assess the quality of translations for any language pair.

\bibliography{custom, anthology}


%

\end{document}